\pdfoutput=1

\documentclass{article}
\usepackage{spconf,amsmath}
\usepackage{tikz}
\usepackage{amssymb, amsfonts}
\usetikzlibrary{matrix}
\usepackage{pgfplots}
\usepackage{colortbl} 
\usepackage{graphicx} 
\usepackage{subcaption}  
\usepackage{booktabs}    
\usepackage{soul}
\usepackage{comment}

\usepackage[top=0.75in, bottom=1.05in, left=0.75in, right=0.75in]{geometry}
\selectfont   
\usepackage[labelsep=period]{caption}
\usepackage{enumitem} 

\usepackage{enumitem}
\setlist{nosep, leftmargin=14pt}
\usepackage{setspace}
\usepackage{mwe} 

\usepackage{marvosym} 

\title{PCA-Enhanced Probabilistic U-Net for Effective Ambiguous Medical Image Segmentation}
\name{\shortstack{
		Xiangyu Li$^{1}$, Chenglin Wang$^{1,2}$, Qiantong Shen$^{3}$,
		Fanding Li$^{1}$, Wei Wang$^{4}$\textsuperscript{\Letter},Kuanquan Wang$^{1}$, Yi Shen$^{1}$, \\
		Baochun Zhao$^{5}$, Gongning Luo$^{1}$ \thanks{\Letter\ Corresponding author: 
			wangwei2019@hit.edu.cn}}
}

\address{$^{1}$ Harbin Institute of Technology, Harbin, China \\
	$^{2}$ Harbin Institute of Technology Zhengzhou Research Institute, China \\
	$^{3}$ Peking University, China \\
	$^{4}$ Harbin Institute of Technology, Shenzhen, China \\
$^{5}$ Hainan College of Software Technology, China}

\begin{document}
%
\maketitle 
\begin{abstract}
Ambiguous Medical Image Segmentation (AMIS) is significant to address the challenges of inherent uncertainties from image ambiguities, noise, and subjective annotations. Existing conditional variational autoencoder (cVAE)-based methods effectively capture uncertainty but face limitations including redundancy in high-dimensional latent spaces and limited expressiveness of single posterior networks. To overcome these issues, we introduce a novel PCA-Enhanced Probabilistic U-Net (\textbf{PEP U-Net}). Our method effectively incorporates Principal Component Analysis (PCA) for dimensionality reduction in the posterior network to mitigate redundancy and improve computational efficiency. Additionally, we further employ an inverse PCA operation to reconstruct critical information, enhancing the latent space's representational capacity. Compared to conventional generative models, our method preserves the ability to generate diverse segmentation hypotheses while achieving a superior balance between segmentation accuracy and predictive variability, thereby advancing the performance of generative modeling in medical image segmentation. 
\end{abstract}

\section{Introduction}
\label{sec:intro}
Deep-learning-based medical image segmentation (MIS) has demonstrated remarkable performance over the past few years \cite{litjens2017survey}. Nonetheless, its clinical adoption faces several challenges, primarily due to uncertainties stemming from inherent image ambiguities. These uncertainties arise from multiple sources, including intrinsic noise, inter-image variability, and the subjective expert annotations. Furthermore, the inherent complexity of anatomical structures and the resolution limitations of medical images introduce additional ambiguity into the analysis. Therefore, it is imperative to address these uncertainties to ensure the reliability of deep learning-based MIS systems. In this context, ambiguous medical image segmentation (AMIS) methods \cite{kohlProbabilisticUNetSegmentation} have emerged, aiming to explicitly model the ambiguity in medical images and thereby forming a cornerstone for trustworthy automated diagnosis. 
\begin{figure}[!t]
	\centering
	\includegraphics[width=0.85\linewidth]{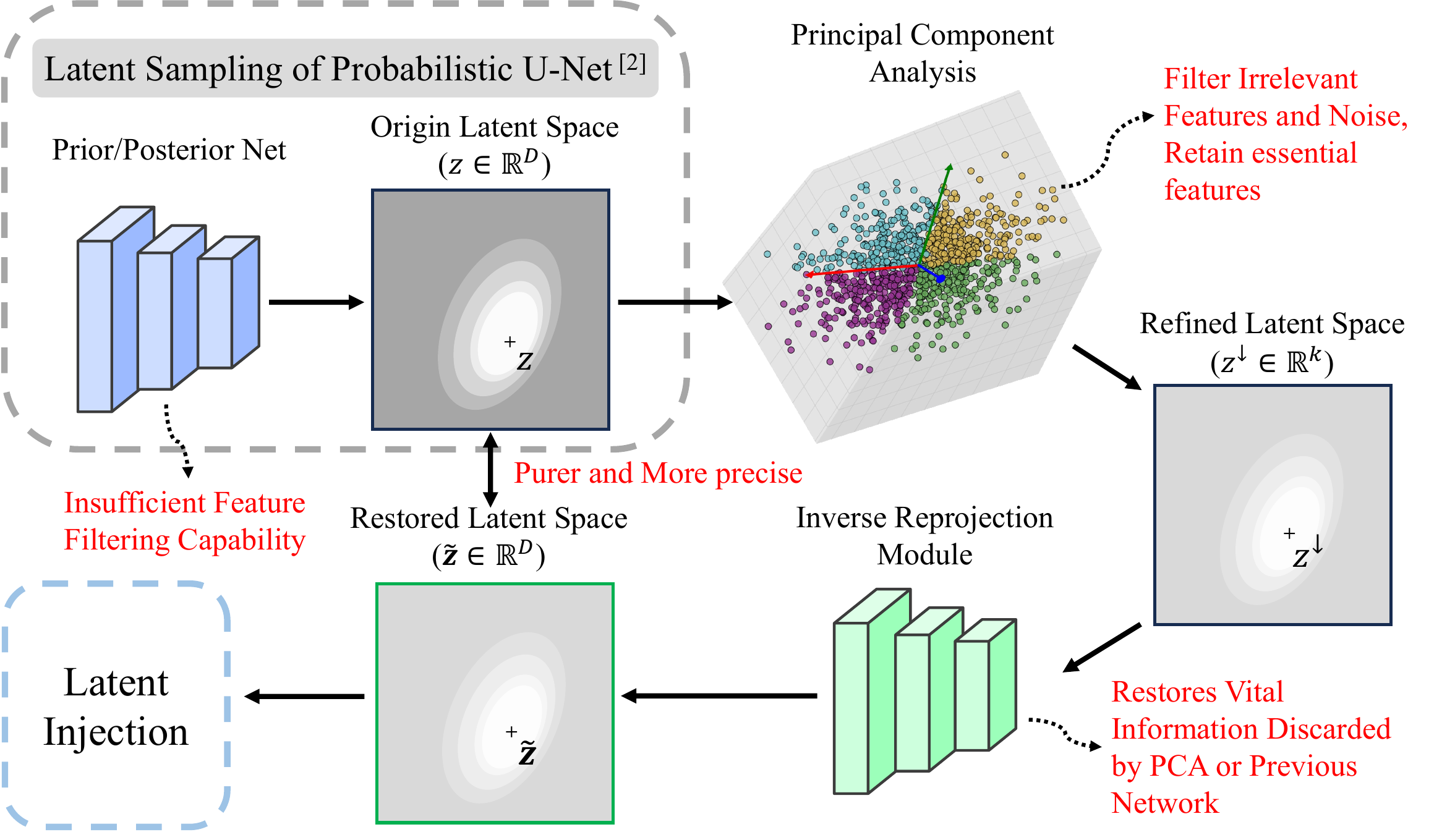}
	\caption{The main principle of the proposed PCA-Enhanced Probabilistic U-Net.}
	\label{fig:intro}
\end{figure}

Over the past few years, significant progress has been made in ambiguous medical image segmentation (AMIS). Early approaches primarily relied on model ensembles \cite{lakshminarayanan2017simple, li2022ultra} or multi-head architectures \cite{Rupprecht_2017_ICCV, li2021hematoma}, which generated multiple segmentation hypotheses by integrating diverse network structures or parallel prediction heads. Although these methods approximate the label distribution to some extent, their outputs often lack semantic correlations.  Consequently, they generally fail to capture a well-defined segmentation distribution, frequently yielding fragmented or inconsistent predictions. More recent studies have introduced stochasticity using conditional variational autoencoders (cVAEs) \cite{kohlProbabilisticUNetSegmentation}. These cVAE-based methods address image ambiguity through Gaussian modeling with dedicated prior and posterior networks. This framework not only produces diverse segmentation samples but also effectively captures predictive uncertainty. 
\begin{figure*}[!t]
	\centering
	\includegraphics[width=0.9\linewidth]{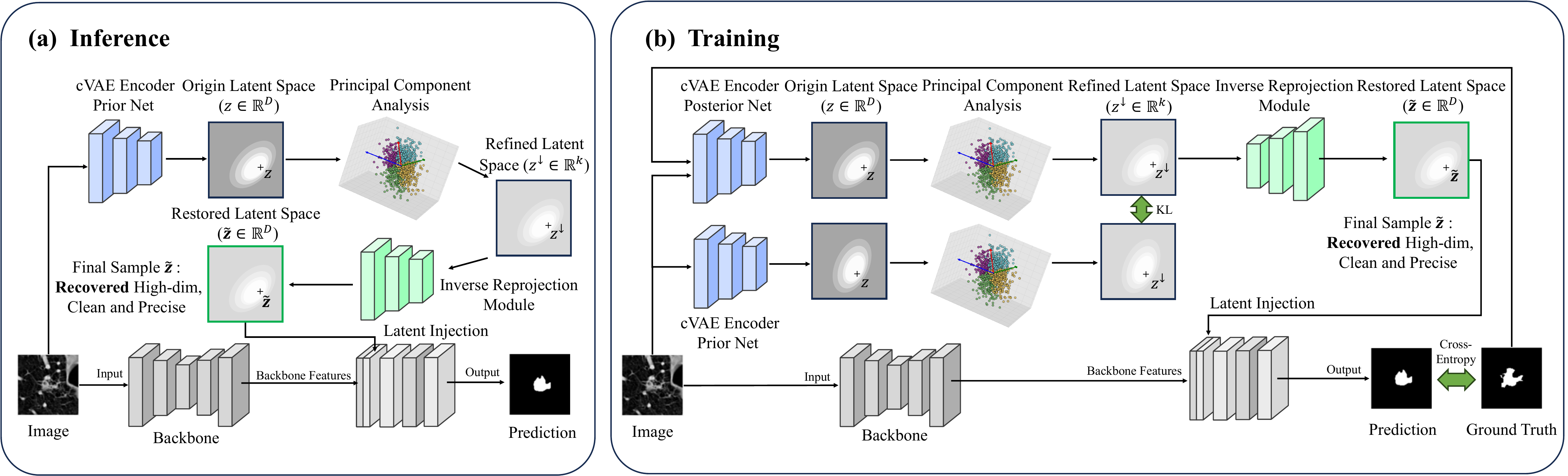}
	\caption{The overall architecture of the proposed PEP U-Net}
	\label{fig:framework}
\end{figure*}

However, existing cVAE-based methods are constrained by several inherent limitations. First, its performance is often dependent on high-dimensional latent spaces. Such high dimensionality can introduce significant redundancy into the learned representations, causing the model's stochasticity to become dispersed and weakening its capacity to capture semantically meaningful uncertain components. This not only increases computational overhead but also complicates the visualization of quantified uncertainties. Second, the architecture employs a single posterior network, which exhibits limited expressiveness in modeling and transforming the latent space. This often results in a pronounced mismatch between the variational approximation used for inference and the true posterior distribution.

To address these limitations, this paper presents an enhanced ambiguous medical image segmentation model named PCA-Enhanced Probabilistic U-Net. Our approach incorporates a dimensionality reduction step using Principal Component Analysis (PCA) \cite{jolliffePrincipalComponentAnalysis2016} in the posterior network to mitigate redundancy in high-dimensional latent spaces, thereby significantly improving computational efficiency. Furthermore, we introduce an inverse PCA operation to reconstruct critical information potentially lost during dimensionality reduction. This reconstruction enhances the latent space's representational capacity and enables more precise uncertainty modeling in medical images. Compared to conventional generative models, the proposed method not only maintains the ability to generate diverse segmentation hypotheses but also achieves a better balance between segmentation accuracy and predictive variability. Consequently, it advances the overall performance of generative modeling in ambiguous medical image segmentation.

\section{Method}\label{sec:method}

\subsection{Overall Framework}
As is illustrated in Fig.\ref{fig:framework}, we proposed a novel PCA-Enhanced Probabilistic U-Net (\textbf{PEP U-Net}) which is built upon several core components. A standard U-Net serves as the deterministic backbone, producing the final segmentation logits. In addition, the proposed model employs a compact latent space, $\mathbb{R}^k$, which is designed to encode plausible segmentation hypotheses. A key innovation lies in our dual-path bottleneck design: both the prior and posterior networks output a Gaussian distribution in a high-dimensional space, $\mathbb{R}^D$, which is then projected onto the same compact $k$-dimensional subspace via a PCA bottleneck. This operation concentrates stochasticity, reduces redundancy, and facilitates efficient sampling and KL divergence alignment during training. Furthermore, to preserve critical information, an inverse reprojection mechanism is incorporated. Before fusion in the decoder, the compact latent sample is projected back to the native $D$-dimensional space and subsequently used to modulate decoder features through a lightweight adapter, such as a $1\times1$ convolution.

\subsection{PCA-enhanced Feature Bottleneck }\label{sec:posterior-pca}

To address the redundancy in high-dimensional latent spaces, we introduce a PCA-enhanced feature bottleneck module after the standard posterior network. This process compresses the posterior distribution into a $k$-dimensional subspace, concentrating variability on the principal uncertainty factors. The KL divergence is computed within this compact space, which reduces computational overhead and enhances interpretability without altering the variational family. The orthonormal projection matrix $U_k$, which forms the top-$k$ principal directions, is derived via PCA on posterior features and shared to project the prior, ensuring KL consistency. This matrix $U_k$, along with a running mean $m$, is kept fixed throughout training. Formally, given an input pair $(X,Y)$, the posterior network outputs: $\mathcal{N}(\mu_{\text{post}},\Sigma_{\text{post}})$ in $\mathbb{R}^D$, which is then projected into the compact posterior distribution:
\begin{equation}
	\mu_{\text{post}}^{(k)} = U_k^{\!\top}(\mu_{\text{post}}-m),\quad
	\Sigma_{\text{post}}^{(k)} = U_k^{\!\top}\Sigma_{\text{post}}U_k,
\end{equation}
yielding $Q^{(k)}(z_k\!\mid\!X,Y)=\mathcal{N}(\mu_{\text{post}}^{(k)},\Sigma_{\text{post}}^{(k)})$ in $\mathbb{R}^k$. For a consistent divergence term, we project the prior by the same mapping:
\begin{equation}
\mu_{\text{prior}}^{(k)} = U_k^{\!\top}(\mu_{\text{prior}}-m),\quad
\Sigma_{\text{prior}}^{(k)} = U_k^{\!\top}\Sigma_{\text{prior}}U_k,
\end{equation}
and compute the KL divergency in the compact space between $Q^{(k)}$ and $P^{(k)}$. Similarly, sampling is performed in this subspace, utilizing $z_k\!\sim\!Q^{(k)}$ during training and $z_k\!\sim\!P^{(k)}$ at test time. This bottleneck effectively eliminates redundancies in the latent representation $z$, concentrating stochasticity on the dominant factors of uncertainty. Consequently, it reduces the sampling cost from $D$ to $k$ dimensions while preserving the Gaussian variational family under linear transformation.

\subsection{Inverse Latent Space Reprojection}\label{sec:inverse-proj}
Although the PCA-based dimensionality reduction mitigates the issues of high dimensionality, the posterior branch remains limited in its expressiveness, hindering the effective ambiguous medical image segmentation for the decoder and resulting in a suboptimal variational match. To overcome this, we introduce \emph{Inverse Latent Space Reprojection}. After sampling a latent vector $z_k$ in the compact $k$-space, we project it back to the native $D$-dimensional space via the inverse linear mapping: 
\begin{equation}
\tilde z \;=\; U_k z_k + m \;\in\; \mathbb{R}^D.
\end{equation}
This differentiable reconstruction restores the rich conditioning capacity needed for effective fusion with the decoder, while the KL divergence and sampling remain efficient in the $k$-space. A lightweight $1{\times}1$ convolutional adapter then transforms $\tilde z$ before it modulates the decoder features. This design ensures compatibility with the standard Probabilistic U-Net, maintains a low-dimensional divergence term, and enables effective conditioning in the native latent space.

\section{Experiment}

\subsection{Datasets}

The \textbf{LIDC dataset \cite{wangHybridUNetbasedDeep2022}} provides comprehensive information on lung nodules, consisting of chest CT scans and corresponding lesion annotations.It contains 1,018 CT scans from 1,010 patients.During the pre-processing stage, images were cropped and normalized the by centering on each lung nodule and extracting patches of size 128 × 128.

\noindent The \textbf{PhC-U373 Dataset\cite{ronnebergerUNetConvolutionalNetworks2015}} contains phase-contrast microscopy images of U373 glioblastoma/astrocytoma cells. For efficient model training, we cropped the images around individual cells, resizing both the raw images and their corresponding labels into 128 × 128 patches.
\subsection{Evaluation Metrics}
In this study, we evaluate segmentation quality using \textbf{Intersection over Union (IoU)} and \textbf{Generalized Energy Distance (GED)}\cite{bellemare2017cramer} metrics. For uncertainty quantification, we evaluate with \textbf{Negative Log Likelihood (NLL)}, \textbf{Brier Score (Br)}\cite{glenn1950verification}, and \textbf{Expected Calibration Error (ECE)}\cite{naeini2015obtaining}, consistent with the criteria employed in~\cite{liHematomaExpansionContext2022}. 

\subsection{Experimental Setup}
All experiments are conducted on an NVIDIA RTX~3090 GPU using the PyTorch framework. 
The network is optimized with the Adam optimizer, where the initial learning rate is set to $1\times10^{-4}$ and no weight decay is applied. 
A StepLR scheduler is employed to decay the learning rate by a factor of $0.1$ every $10$ epochs, and the maximum number of training epochs is set to $200$. 
For each dataset, $20\%$ of the samples are randomly selected for testing, while the remaining $80\%$ are used for training. 
The batch size is set to $8$ during training and $1$ during inference. 
The latent dimensionality of the original latent space is set to $D=6$, which is subsequently projected to a compact subspace of $k=2$ for LIDC and $k=3$ for PhC dimensions via PCA, providing a favorable balance between multi-modality expressiveness and computational efficiency. 

\begin{table}[!t]
	\centering
	\caption{Results of different models on the LIDC and PhC datasets}
	\label{tab:combined-results}
	\small
	\begin{tabular}{lcccc}
		\toprule
		& \multicolumn{2}{c}{LIDC dataset} & \multicolumn{2}{c}{PhC dataset} \\
		\cmidrule(lr){2-3} \cmidrule(lr){4-5}
		Model & IoU $\uparrow$ & GED $\downarrow$ & IoU $\uparrow$ & GED $\downarrow$ \\
		\midrule
		U-Net & 0.322 & -- & 0.878 & -- \\
		Prob. U-Net\cite{kohlProbabilisticUNetSegmentation} & 0.397 & 0.180 & 0.882 & 0.019 \\
		SSN \cite{monteiro2020stochastic} & 0.412 & 0.220 & 0.900 & 0.020 \\
		cSSN \cite{zepf2023label} & 0.419 & 0.241 & \textbf{0.908} & 0.026 \\
		\hline
		PEP U-Net w/o ILSR  & 0.407 & 0.160 & 0.883 & 0.010 \\
		PEP U-Net & \textbf{0.434} & \textbf{0.120} & 0.890 & \textbf{0.008} \\
		\bottomrule
	\end{tabular}
\end{table}
\begin{table*}[htbp]
	\centering
	\small
	\caption{Uncertainty estimation metrics on the LIDC and PhC datasets.}
	\label{tab:abl-combined}
	\begin{tabular}{lcccccc}
		\toprule
		& \multicolumn{3}{c}{LIDC dataset} & \multicolumn{3}{c}{PhC dataset} \\
		\cmidrule(lr){2-4} \cmidrule(lr){5-7}
		Model & NLL $\downarrow$ & Brier (\%) $\downarrow$ & ECE (\%) $\downarrow$ & NLL $\downarrow$ & Brier (\%) $\downarrow$ & ECE (\%) $\downarrow$ \\
		\midrule
		Prob. U-Net\cite{kohlProbabilisticUNetSegmentation} & 0.013 & 0.35 & 0.18 & 0.087 & 2.530 & 1.10 \\
		SSN\cite{monteiro2020stochastic} & 0.012 & \textbf{0.102} & 0.235 & - & \textbf{0.634} & 1.867 \\
		c-SSN\cite{zepf2023label} & - & - & - & - & 1.109 & 2.713 \\
		\hline
		PEP U-Net w/o ILSR & 0.012 & 0.310 & 0.130 & 0.086 & 2.520 & 0.820 \\
		PEP U-Net & \textbf{0.009} & 0.280 & \textbf{0.070} & \textbf{0.085} & 2.500 & \textbf{0.620} \\
		\bottomrule
	\end{tabular}
\end{table*}
\subsection{Results and Discussions}
In the comparative experiments, the proposed PEP U-Net was evaluated against several baseline and state-of-the-art approaches, including U-Net, Probabilistic U-Net\cite{kohlProbabilisticUNetSegmentation}, Stochastic Segmentation Network (SSN)\cite{monteiro2020stochastic}, and its variant cSSN\cite{zepf2023label}. The experimental results demonstrate that our method achieves a superior balance between segmentation accuracy and uncertainty estimation reliability. As shown in Table \ref{tab:combined-results}, on the challenging LIDC dataset, PEP U-Net achieves the highest IoU of 0.434 and the lowest GED of 0.120, significantly outperforming all comparative methods. While cSSN shows competitive performance on the PhC dataset with an IoU of 0.908, PEP U-Net maintains strong performance with an IoU of 0.890 and achieves the best GED of 0.008, indicating superior ambiguous medical image segmentation capability. Further analysis of uncertainty estimation metrics in Table \ref{tab:abl-combined} reveals that PEP U-Net consistently delivers excellent performance across multiple calibration metrics. On the LIDC dataset, it achieves the lowest NLL (0.009) and ECE (0.07\%), while on the PhC dataset, it obtains the best NLL (0.086) and ECE (0.62\%). Although SSN shows competitive Brier scores on both datasets, PEP U-Net maintains well-balanced performance across all uncertainty metrics. 
The consistent improvement across both accuracy-oriented and calibration-focused metrics underscores the effectiveness of our method in producing trustworthy predictions for medical image segmentation tasks. 
\begin{figure}[htbp]
	\centering
	\includegraphics[width=\linewidth]{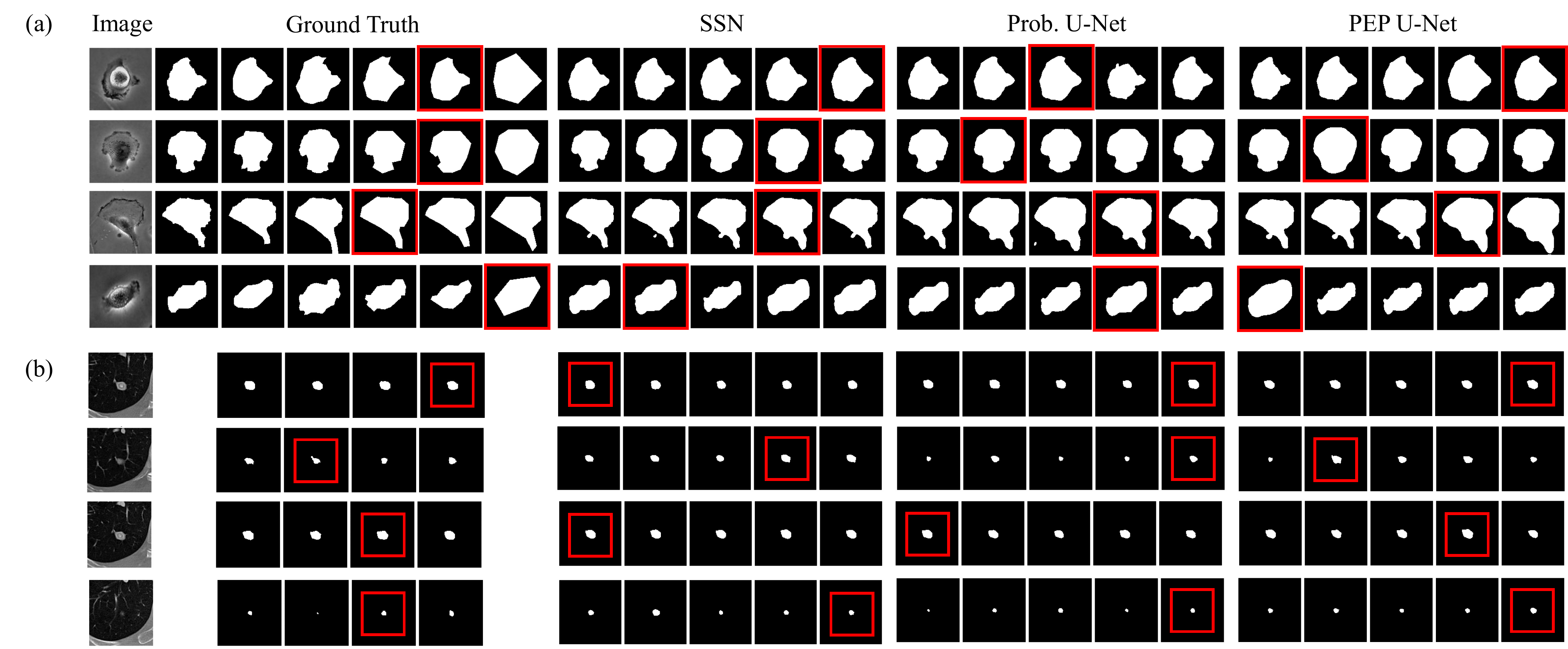}
	\caption{Qualitative experimental results on LIDC and PhC datasets}
	\label{fig:visual-res}
\end{figure}
\subsection{Ablation Studies}
To evaluate the contribution of each proposed component, we conducted a comprehensive ablation study. The model excluding the inverse latent space reprojection (ILSR)—is denoted as \textbf{PEP U-Net w/o ILSR}, and the baseline model without these two modules as \textbf{Prob. U-Net}. As summarized in Tables~\ref{tab:combined-results} and~\ref{tab:abl-combined}, both the PCA-based dimensionality reduction and the ILSR operation contribute notably to improve segmentation accuracy and uncertainty modeling. Quantitative results reveal a consistent performance hierarchy across datasets: the full PEP U-Net achieves the best results, followed by PEP U-Net w/o ILSR, with the original Probabilistic U-Net yielding the lowest performance among the three. This trend is reflected not only in segmentation metrics such as IoU and GED but also in uncertainty quantification measures including NLL, Brier score, and ECE. 
\subsection{Hyperparameter Experiments for PCA-Retained Dimensions}

In this study, the hyperparameter PCA-Retained Dimensions ($k$) is shown to significantly influence both the accuracy and diversity of segmentation outcomes. To identify its optimal value, we conducted a supplementary experiment in which $k$ was systematically varied and evaluated in terms of IoU and GED on the test set. Beyond hyperparameter tuning, this experiment also provides valuable insights into the behavioral patterns of the proposed model under different latent space dimensions. The corresponding results are summarized in Table~\ref{tab:hp-results}. As shown in the tables, the number of principal components has a noticeable impact on the performance of the PEP U-Net across different datasets. For the LIDC dataset, the model performs best when the number of principal components is set to $k=2$. As the number increases, redundant information may be introduced, which adversely affects the model's segmentation capability. In contrast, for the PhC dataset, the optimal performance is achieved with $k=3$ principal components. These findings suggest that the optimal number of principal components varies depending on the dataset, which can be attributed to differences in the inherent characteristics of each dataset. 
\begin{table}[htbp]
	\centering
	\small  
	\caption{Performance test results of the PEP U-Net model retaining different principal components.}
	\label{tab:hp-results}
	\begin{tabular}{lcccc}
		\toprule
		& \multicolumn{2}{c}{LIDC dataset} & \multicolumn{2}{c}{PhC dataset} \\
		\cmidrule(lr){2-3} \cmidrule(lr){4-5}
		$k$ & IoU $\uparrow$ & GED $\downarrow$ & IoU $\uparrow$ & GED $\downarrow$ \\
		\midrule
		2 & \textbf{0.434} & \textbf{0.120} & 0.884 & 0.009 \\
		3 & 0.416 & 0.146 & \textbf{0.890} & \textbf{0.008} \\
		4 & 0.397 & 0.152 & 0.875 & 0.009 \\
		\bottomrule
	\end{tabular}
\end{table}
\section{Conclusion}
In this work, we have addressed key limitations in existing conditional variational autoencoder-based ambiguous medical image segmentation methods by proposing an enhanced Probabilistic U-Net framework. Our approach systematically tackles the problems of redundancy in high-dimensional latent spaces and the limited expressiveness of single posterior networks through the integration of Principal Component Analysis. The incorporation of PCA-based dimensionality reduction in the posterior network effectively mitigates feature redundancy while significantly improving computational efficiency. Furthermore, the innovative inverse PCA operation enables reconstruction of critical information that might otherwise be lost during dimensionality reduction, thereby enhancing the representational capacity of the latent space. Experimental results demonstrate that our method maintains the crucial ability to generate diverse segmentation hypotheses while achieving a superior balance between segmentation accuracy and predictive variability compared to conventional generative models. 

\section {Compliance with Ethical Standards}
This research study was conducted retrospectively using human subject data made available in open access by (LIDC and PhC). Ethical approval was not required as confirmed by the license attached with the open access data.
\section{ACKNOWLEDGMENTS}
This work was supported by the National Natural Science Foundation of China under Grants 62501195, 62272135, 62372135, and the Key Research \& Development Program of Heilongjiang Province under Grant 2024ZX12C23, and the Natural Science Foundation of Heilongjiang Province under Grants LH2024F019.

\begin{thebibliography}{10}

\bibitem{litjens2017survey}
Geert Litjens, Thijs Kooi, Babak~Ehteshami Bejnordi, Arnaud Arindra~Adiyoso
  Setio, Francesco Ciompi, Mohsen Ghafoorian, Jeroen~Awm Van Der~Laak, Bram
  Van~Ginneken, and Clara~I S{\'a}nchez,
\newblock ``A survey on deep learning in medical image analysis,''
\newblock {\em Medical image analysis}, vol. 42, pp. 60--88, 2017.

\bibitem{kohlProbabilisticUNetSegmentation}
Simon Kohl, Bernardino {Romera-Paredes}, Clemens Meyer, Jeffrey~De Fauw,
  Joseph~R Ledsam, Klaus {Maier-Hein}, S~M~Ali Eslami, Danilo~Jimenez Rezende,
  and Olaf Ronneberger,
\newblock ``A {{Probabilistic U-Net}} for {{Segmentation}} of {{Ambiguous
  Images}},''
\newblock .

\bibitem{lakshminarayanan2017simple}
Balaji Lakshminarayanan, Alexander Pritzel, and Charles Blundell,
\newblock ``Simple and scalable predictive uncertainty estimation using deep
  ensembles,''
\newblock {\em Advances in neural information processing systems}, vol. 30,
  2017.

\bibitem{li2022ultra}
Xiangyu Li, Xinjie Liang, Gongning Luo, Wei Wang, Kuanquan Wang, and Shuo Li,
\newblock ``Ultra: Uncertainty-aware label distribution learning for breast
  tumor cellularity assessment,''
\newblock in {\em Medical Image Computing and Computer Assisted
  Intervention--MICCAI 2022: 25th International Conference, Singapore,
  September 18--22, 2022, Proceedings, Part III}. Springer, 2022, pp. 303--312.

\bibitem{Rupprecht_2017_ICCV}
Christian Rupprecht, Iro Laina, Robert DiPietro, Maximilian Baust, Federico
  Tombari, Nassir Navab, and Gregory~D. Hager,
\newblock ``Learning in an uncertain world: Representing ambiguity through
  multiple hypotheses,''
\newblock in {\em Proceedings of the IEEE International Conference on Computer
  Vision (ICCV)}, Oct 2017.

\bibitem{li2021hematoma}
Xiangyu Li, Gongning Luo, Wei Wang, Kuanquan Wang, Yue Gao, and Shuo Li,
\newblock ``Hematoma expansion context guided intracranial hemorrhage
  segmentation and uncertainty estimation,''
\newblock {\em IEEE Journal of Biomedical and Health Informatics}, vol. 26, no.
  3, pp. 1140--1151, 2021.

\bibitem{jolliffePrincipalComponentAnalysis2016}
Ian~T. Jolliffe and Jorge Cadima,
\newblock ``Principal component analysis: A review and recent developments,''
\newblock {\em Philosophical Transactions of the Royal Society A: Mathematical,
  Physical and Engineering Sciences}, vol. 374, no. 2065, pp. 20150202, Apr.
  2016.

\bibitem{wangHybridUNetbasedDeep2022}
Yifan Wang, Chuan Zhou, Heang-Ping Chan, Lubomir~M. Hadjiiski, Aamer Chughtai,
  and Ella~A. Kazerooni,
\newblock ``Hybrid {{U}}-{{Net}}-based deep learning model for volume
  segmentation of lung nodules in {{CT}} images,''
\newblock {\em Medical Physics}, vol. 49, no. 11, pp. 7287--7302, Nov. 2022.

\bibitem{ronnebergerUNetConvolutionalNetworks2015}
Olaf Ronneberger, Philipp Fischer, and Thomas Brox,
\newblock ``U-{{Net}}: {{Convolutional Networks}} for {{Biomedical Image
  Segmentation}},''
\newblock in {\em Medical {{Image Computing}} and {{Computer-Assisted
  Intervention}} -- {{MICCAI}} 2015}, Nassir Navab, Joachim Hornegger,
  William~M. Wells, and Alejandro~F. Frangi, Eds., vol. 9351, pp. 234--241.
  Springer International Publishing, Cham, 2015.

\bibitem{bellemare2017cramer}
Marc~G Bellemare, Ivo Danihelka, Will Dabney, Shakir Mohamed, Balaji
  Lakshminarayanan, Stephan Hoyer, and R{\'e}mi Munos,
\newblock ``The cramer distance as a solution to biased wasserstein
  gradients,''
\newblock {\em arXiv preprint arXiv:1705.10743}, 2017.

\bibitem{glenn1950verification}
W~Brier Glenn et~al.,
\newblock ``Verification of forecasts expressed in terms of probability,''
\newblock {\em Monthly weather review}, vol. 78, no. 1, pp. 1--3, 1950.

\bibitem{naeini2015obtaining}
Mahdi~Pakdaman Naeini, Gregory Cooper, and Milos Hauskrecht,
\newblock ``Obtaining well calibrated probabilities using bayesian binning,''
\newblock in {\em Proceedings of the AAAI conference on artificial
  intelligence}, 2015, vol.~29.

\bibitem{liHematomaExpansionContext2022}
Xiangyu Li, Gongning Luo, Wei Wang, Kuanquan Wang, Yue Gao, and Shuo Li,
\newblock ``Hematoma {{Expansion Context Guided Intracranial Hemorrhage
  Segmentation}} and {{Uncertainty Estimation}},''
\newblock {\em IEEE Journal of Biomedical and Health Informatics}, vol. 26, no.
  3, pp. 1140--1151, Mar. 2022.

\bibitem{monteiro2020stochastic}
Miguel Monteiro, Lo{\"\i}c Le~Folgoc, Daniel Coelho~de Castro, Nick Pawlowski,
  Bernardo Marques, Konstantinos Kamnitsas, Mark Van~der Wilk, and Ben Glocker,
\newblock ``Stochastic segmentation networks: Modelling spatially correlated
  aleatoric uncertainty,''
\newblock {\em Advances in neural information processing systems}, vol. 33, pp.
  12756--12767, 2020.

\bibitem{zepf2023label}
Kilian Zepf, Eike Petersen, Jes Frellsen, and Aasa Feragen,
\newblock ``That label's got style: Handling label style bias for uncertain
  image segmentation,''
\newblock {\em arXiv preprint arXiv:2303.15850}, 2023.

\end{thebibliography}

\end{document}